\documentclass[final]{cvpr}

\usepackage{times}
\usepackage{epsfig}
\usepackage{graphicx}
\usepackage{amsmath}
\usepackage{amssymb}
\usepackage{subcaption}

\usepackage[pagebackref=true,breaklinks=true,colorlinks,bookmarks=false]{hyperref}

\def\cvprPaperID{3106} 
\def\confYear{CVPR 2021}

\begin{document}

\title{CT Film Recovery via Disentangling Geometric Deformation and Illumination Variation: Simulated Datasets and Deep Models}

\author{Quan Quan\\
Institute of Computing Technology, CAS\\
{\tt\small quan.quan@miracle.ict.ac.cn}
\and
Qiyuan Wang\\
Nanjing University\\
{\tt\small MF1823053@smail.nju.edu.cn}
\and
Liu Li\\
Imperial College London\\
{\tt\small liu.li20@imperial.ac.uk}
\and
Yuanqi Du\\
George Mason University\\
{\tt\small ydu6@gmu.edu}
\and
S. Kevin Zhou\\
Institute of Computing Technology, CAS\\
{\tt\small s.kevin.zhou@gmail.com}
}

\maketitle

\begin{abstract}

While medical images such as computed tomography (CT) are stored in DICOM format in hospital PACS, it is still quite routine in many countries to print a film as a transferable medium for the purposes of self-storage and secondary consultation. Also, with the ubiquitousness of mobile phone cameras, it is quite common to take pictures of the CT films, which unfortunately suffer from geometric deformation and illumination variation. In this work, we study the problem of recovering a CT film, which marks the first attempt in the literature, to the best of our knowledge. We start with building a large-scale head CT film database CTFilm20K, consisting of approximately 20,000 pictures, using the widely used computer graphics software Blender. We also record all accompanying information related to the geometric deformation (such as 3D coordinate, depth, normal, and UV maps) and illumination variation (such as albedo map). Then we propose a deep framework to disentangle geometric deformation and illumination variation using the multiple maps extracted from the CT films to collaboratively guide the recovery process. Extensive experiments on simulated and real images demonstrate the superiority of our approach over the previous approaches. We plan to open source the simulated images and deep models for promoting the research on CT film recovery\footnote{https://anonymous.4open.science/r/e6b1f6e3-9b36-423f-a225-55b7d0b55523/}. 
\end{abstract}

\section{Introduction}

    Recently, more and more attention has been drawn to medical image analysis\cite{zhou2019handbook}. For example, the spreading of COVID-19 pandemic greatly promotes the use of computed tomography (CT) in clinical practice. While medical images such as CT are stored in DICOM format in hospital PACS, it is still quite routine in many countries to print a film as a transferable medium for the purposes of self-storage and secondary consultation. For example, the market size of global medical radiography film, which is commonly used to print a stack of CT slices, is estimated to reach 986.1 million USD by 2026, at a CAGR of 0.4\% since 2016\footnote{https://www.marketwatch.com/press-release/medical-x-ray-film-market-size-share-global-trends-market-demand-industry-analysis-growth-opportunities-and-forecast-2026-2020-11-03}. 

    \begin{figure}[t] 
    \begin{center}
       \includegraphics[width=\linewidth]{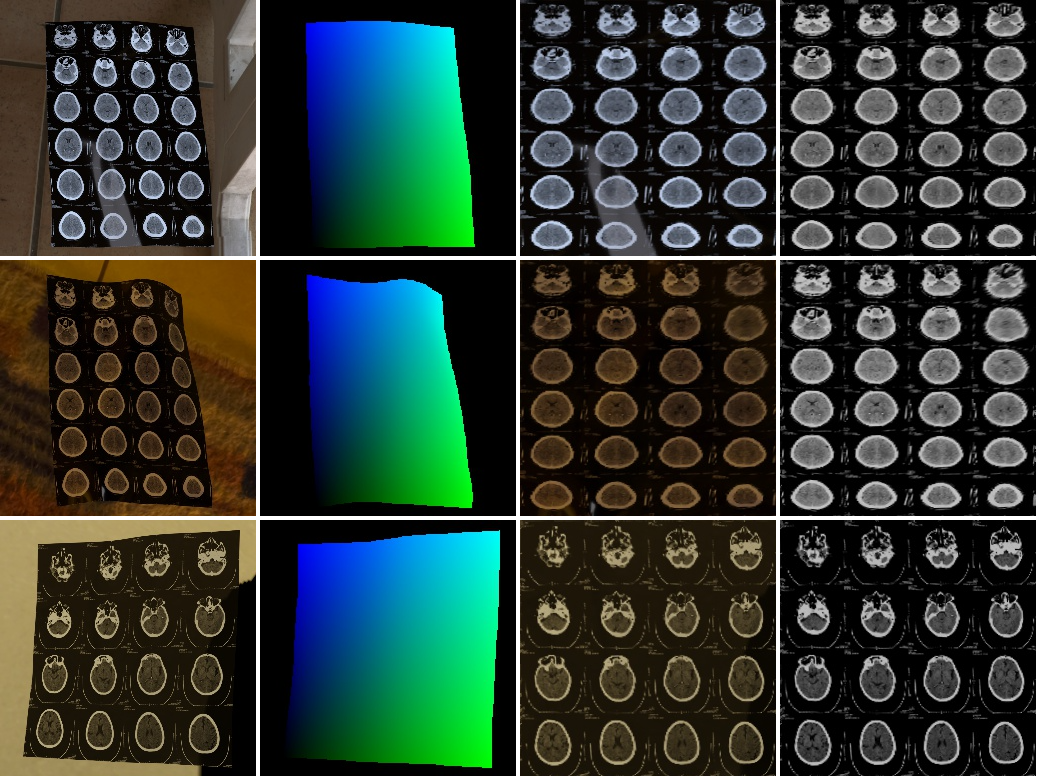}
    \end{center}
       \caption{\textbf{CT film recovery.} The 1st column: warped CT films. The 2nd column: UV map. The 3rd column: dewarped films. The 4th column: de-illuminated \& dewarped films.}
       \label{fig:demo1}
    \end{figure}
    
    With the ubiquitousness of mobile phone cameras, it becomes quite common to take pictures of the CT films to avoid physically transferring CT films or support remote access. 
    While convenient, the pictures most time do not meet the standard for the doctors to process and analyze because sometimes the CT films are warped in the pictures.
    In addition to the geometric deformation, there is illumination variation too as the lighting condition and background scene in which the film is presented are rather unconstrained. Therefore, it is a challenge to both dewarp the pictures and recover the content from rectified CT films at the same time. In this paper, we make the first attempt in the literature, to the best of our knowledge, to propose a CT film recovery solution. Its crucial part is to remove the background in the image and rectify it as in Figure~\ref{fig:demo1}.

    \begin{figure}[t] 
    \begin{center}
       \includegraphics[width=\linewidth]{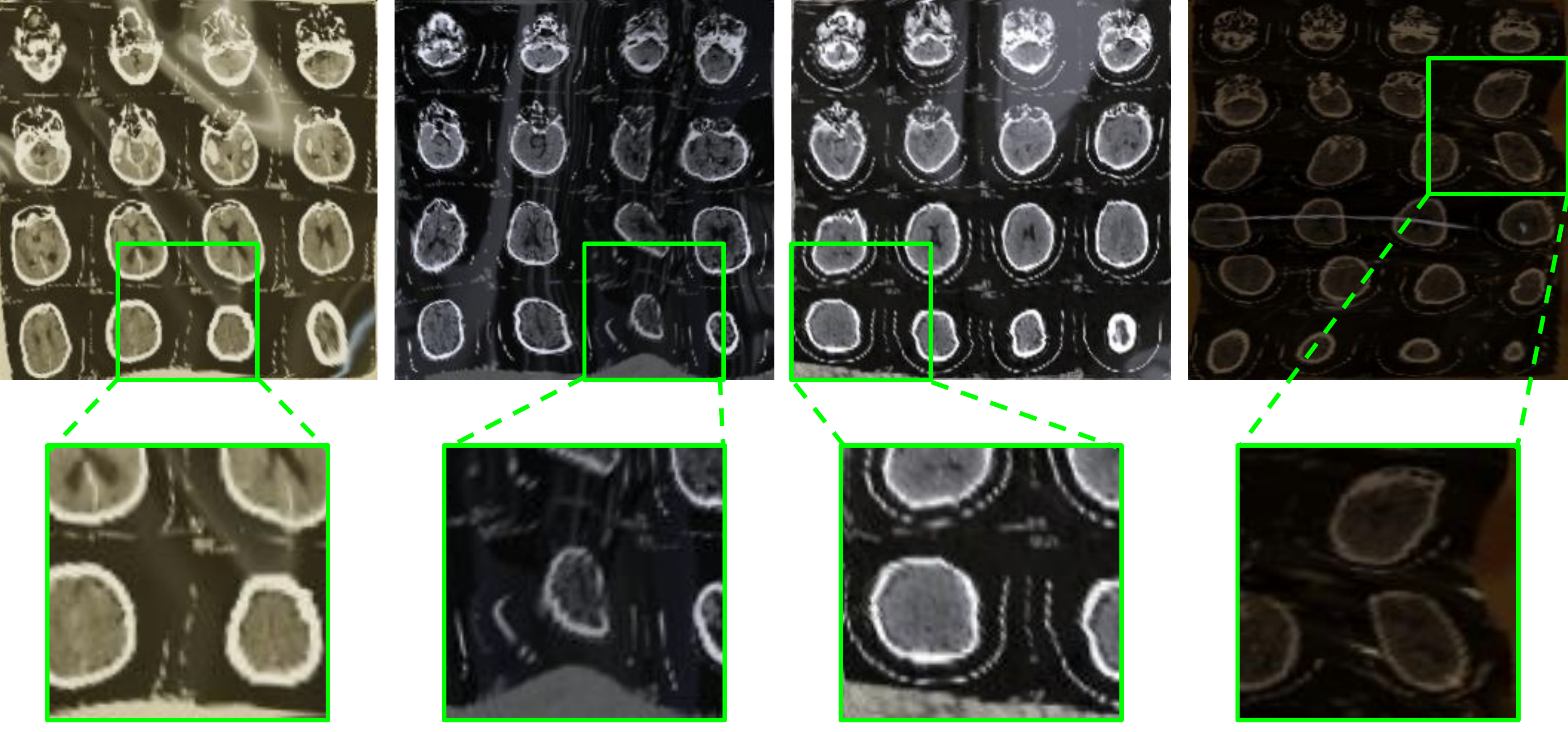}
    \end{center}
       \caption{\textbf{Examples of warped regions.} Top row: recovered images from the baseline model of DewarpNet. Bottom row: the zoom-in warped regions of green boxes. }
       \label{fig:warps}
    \end{figure}
    
    Although there are no existing works about recovering warped CT films previously, a more related problem, the recovery and de-distortion of document images, has been studied. The previous methods usually rely on the geometric properties of papers to recover the document~\cite{meng2018exploiting, zhang2009unified}.
    However, these methods are resource- and time-intensive. Later, with the advancement of deep learning on computer vision, researchers start to tackle this problem utilizing the convolutional neural network~\cite{ronneberger2015u}. Recently, DocUNet~\cite{ma2018docunet} firstly tackle this problem by constructing and training large 2D document datasets. DewarpNet~\cite{das2019dewarpnet} follows ~\cite{ma2018docunet} to construct a 3D document dataset and uses the 3D coordinate map, which contains more information about the spatial structure of the documents, to generate a backward map, which is used to restore the warped image in the final step. However, in our CT film recovery task, as in Figure~\ref{fig:warps}, the backward map will cause many locally-warped areas, leading to poor visual performance.
    These locally-warped regions can negatively affect our further medical image processing and clinical analysis for CT films, hence we find another way to obtain the dewarped image\textemdash using a UV map. Though a UV map is generally applied to map textures to models, it can be also used to obtain textures from \textquotedblleft CT film models \textquotedblright, and is verified to generate less locally-warped regions.
    Another problem is that the ambient light mixed with various colors reduce the quality of CT film images, and is harmful to the accurate diagnosis. Therefore, we need to remove the disturbance of lights and focus more on image contrast and clarity. 
    
    Based on these ideas, we propose a \textbf{Film Recovering Network} (abbreviated as \textbf{FiReNet}) that recovers CT film by the UV map which is more robust than the backward map in \cite{das2019dewarpnet}. The FiReNet includes three modules: 1) \textbf{Multi-map module} is designed to improve the quality of UV maps by outputting and combining multiple annotation maps. 2) \textbf{Transformation module} is designed to generate UV maps, and also deformation maps for enhancing the UV map. In addition, to focus on disentangling local warps, both maps are trained with additional customized loss. 3) \textbf{De-illumination module} is designed to solve the illumination problem by outputting albedo maps which retain only the content of the film without the light information. Finally, the albedo map is converted to the unwarped film by using the backward map generated by the FiReNet.

    As three are no public training data for CT film recovery, we construct our \textbf{CTFilm20K Dataset}, which contains a large number of CT films with various real-world warping scenarios and different contents. After collecting high-quality background images and CT data, we utilize photorealistic rendering to generate images. Each sample contains several labeled maps including 3D coordinate, normal, UV, and albedo maps. In total, the dataset contains about 20,000 richly annotated high-quality film images.
    
    Our contributions are summarized as follows:
    \begin{itemize}
        \item We construct the CTFilm20K dataset, which is, to the best of our knowledge, the first CT film repository with multiple ground-truth annotations for collaborative modeling.
        \item We propose FiReNet, a CT film recovery network that benefits from UV maps enhanced by deformation maps and intermediate features from multiple annotation maps and enables robust and high-quality CT film dewarping and de-illumination.
        \item We conduct extensive experiments that train the proposed FiReNet with our CT Film dataset and the model performs better than the previous competing methods both quantitatively and visually.
    \end{itemize}

\section{Previous Work}
There is little work on CT film recovery, while a more related research topic to this problem is the restoration and de-distortion of document images. We classify the related methods into two categories: shape from visual cues and 3D shape reconstruction.

\subsection{Shape from Visual Cues}
Visual cues include lighting/shading, text lines, etc. Wada et al.~\cite{wada1997shape} address such a problem that shadows of warped documents vary under the same directional light sources, and construct shapes by using this phenomenon. Courteille et al.~\cite{courteille2007shape} follow this work with cameras instead of scanners. Zhang et al.~\cite{zhang2009unified} improve the SfS system proposed in \cite{courteille2007shape} to handle both shadows and background noise. Some methods find another way of using the content information to predict the shape of the document. One of the most common strategies is to track the text line~\cite{ezaki2005dewarping, ulges2005document, lu2006document, kim2015document, liu2015restoring}. Cao et al.~\cite{cao2003cylindrical} propose to simulate a warped document with the aid of a cylinder for the first time. Liang et al.~\cite{liang2008geometric} use a developable surface. Tian and Narasimhan~\cite{tian2011rectification} further optimize the text lines as horizontal cues and character strokes as vertical cues to construct a 3D grid. These works can be also seen as special cases of the shape from texture problem~\cite{witkin1981recovering, malik1997computing, forsyth2001shape}. Recently, Das et al.~\cite{das2017common} use CNN to check the correction by detecting paper creases. However, CNN is only used as a tool in the optimization process and is not trained in an end-to-end manner.

\subsection{3D Shape Reconstruction}
There are many ways of reconstructing the 3D shape of paper documents. Brown and Seales~\cite{brown2001document} build a visible light projector-camera system. Zhang et al.~\cite{zhang2008improved} restore the shape by the physical properties of the paper with a more advanced range/depth sensor. Tsoi and Brown~\cite{tsoi2007multi} take the advantage of the boundary information from multi-view images and combine them to generate a rectified image. Similarly, Koo et al.~\cite{koo2009composition} use two images captured from different viewing angles to predict 3D shape by SIFT matching. Meng et al.~\cite{meng2014active} obtain document curling by using two structured laser beams. Ostlund et al.~\cite{ostlund2012laplacian} reconstruct the 3D shape of a deformable surface through the related document image.
Ulges et al.~\cite{ulges2004document} predict the disparity map between the two images by image patch matching. Yamashita et al.~\cite{yamashita2004shape} utilize the non-uniform rational B-spline curve (NURBS) to parameterize the 3D shape of the document. You et al.~\cite{you2017multiview} reconstruct the 3D shape by simulating the creases on multiple document images. Recently, DocUNet~\cite{ma2018docunet} and DewarpNet~\cite{das2019dewarpnet} adopt a data-driven approach to tackle document unwarping with deep learning. In particular, DewarpNet uses the 3D map and normal map as labeled information, and the method of generating backward mapping is not point-aligned, and it does not make full use of multiple labeled images. 

\section{CT Film Dataset: CTFilm20K}
\subsection{Overview of Dataset}
\begin{figure}[t]
    \begin{center}
      \includegraphics[width=\linewidth]{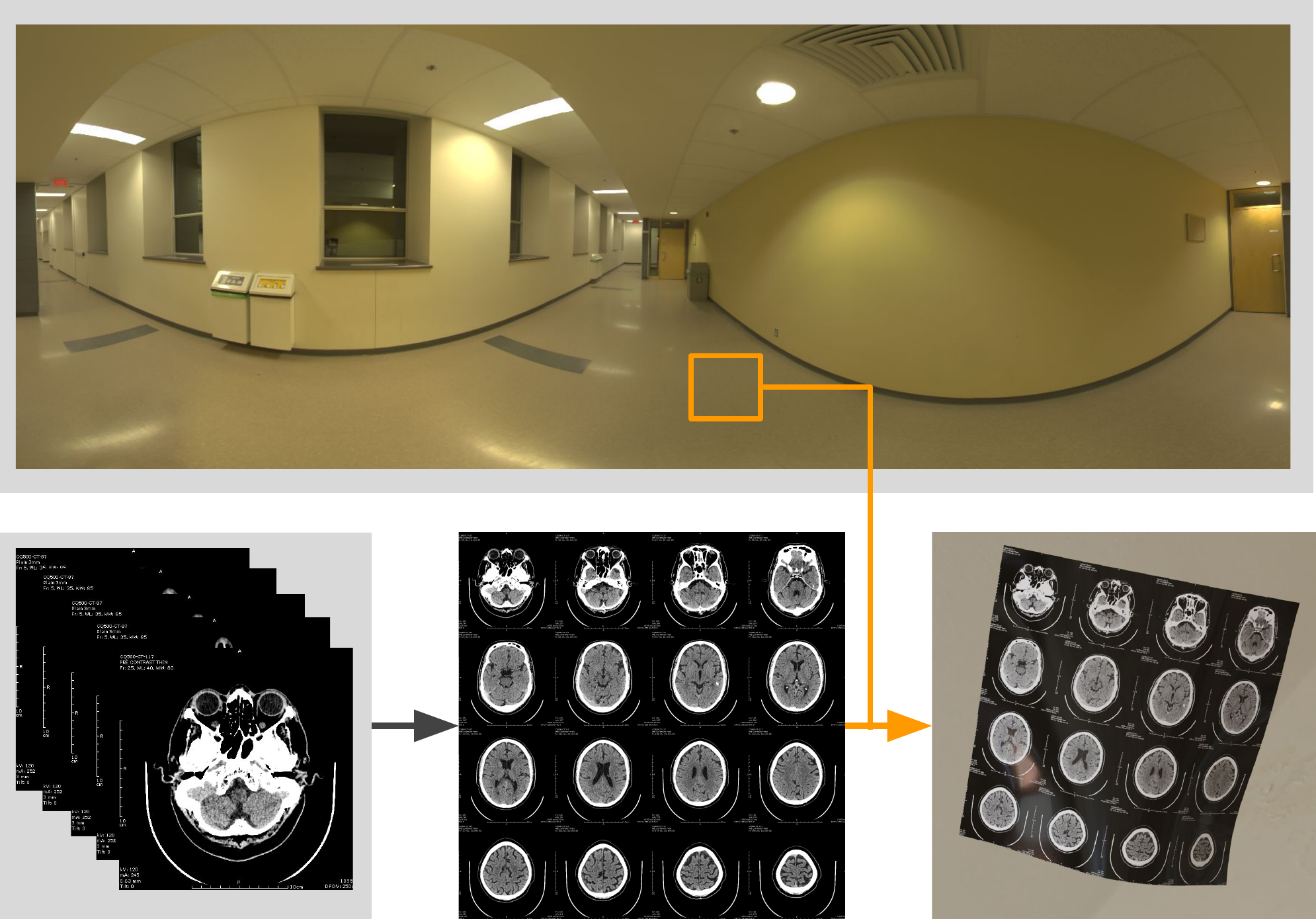}
    \end{center}
      \caption{\textbf{CT film data construction.} Top: environment images. Bottom-left: a series of CT slices. Bottom-middle: CT films from joint slices. Bottom-right: warped CT films. Black arrow: arranging CT slices to a CT film. Yellow arrow: rendering warped CT films with Blender\cite{blender}}
        \label{fig:dataset2}
    \end{figure}

Considering the hard-obtainable characteristic of the medical film data, we create this dataset by computer graphics rendering (Figure~\ref{fig:dataset2}). We first extract slices from the public medical CT DICOMs~\cite{chilamkurthy2018development} and convert them in the grid format to generate the film texture. In comparison with Doc3D~\cite{das2019dewarpnet}, which captures the 3D shape of naturally deformed real objects, we rely on the embedded physics computing engines in Blender~\cite{blender}, a rendering software, which could simulate deformation for different situations. After that, we render the images with synthetic film textures laid in various HDR scenes from~\cite{DBLP:journals/tog/GardnerSYSGGL17} and HDR Haven\footnote{HDRIHAVEN: https://hdrihaven.com/ \label{hdrhaven}} using path tracing. We randomly set camera positions in a certain range with the child-of relationship constraints to film objects and vary illumination conditions in rendering.

\begin{figure}
\centering
\begin{subfigure}{.11\textwidth}
  \centering
  \includegraphics[width=1\linewidth]{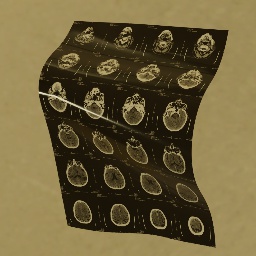}  
  \caption{CT film}
\end{subfigure}
\begin{subfigure}{.11\textwidth}
  \centering
  \includegraphics[width=1\linewidth]{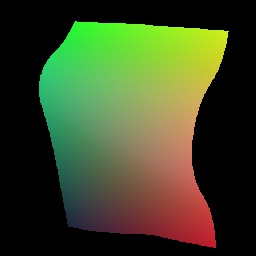}  
  \caption{3D}
\end{subfigure}
\begin{subfigure}{.11\textwidth}
  \centering
  \includegraphics[width=1\linewidth]{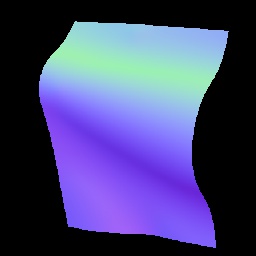}  
  \caption{Normal}
\end{subfigure}
\begin{subfigure}{.11\textwidth}
  \centering
  \includegraphics[width=1\linewidth]{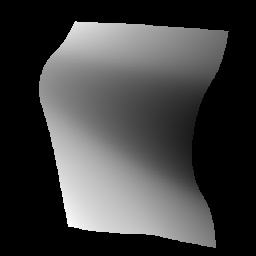}  
  \caption{Depth}
\end{subfigure}

\begin{subfigure}{.11\textwidth}
  \centering
  \includegraphics[width=1\linewidth]{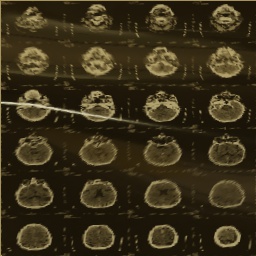}  
  \caption{GT}
\end{subfigure}
\begin{subfigure}{.11\textwidth}
  \centering
  \includegraphics[width=1\linewidth]{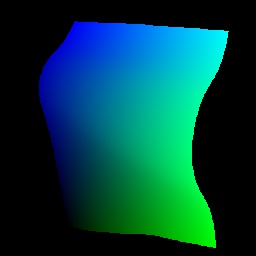}  
  \caption{UV}
\end{subfigure}
\begin{subfigure}{.11\textwidth}
  \centering
  \includegraphics[width=1\linewidth]{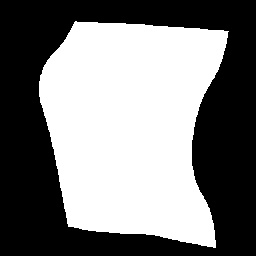}  
  \caption{BG}
\end{subfigure}
\begin{subfigure}{.11\textwidth}
  \centering
  \includegraphics[width=1\linewidth]{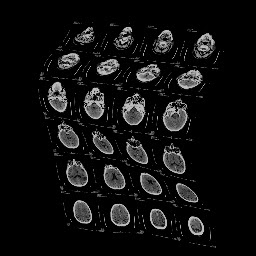}  
  \caption{Albedo}
\end{subfigure}
\caption{\textbf{Components in CT film dataset.} (a) Warped CT film. (b) 3D coordinate map. (c) Normal map. (d) Depth map. (e) Dewarped film. (f) UV map. (g) Background mask. (h) Albedo map.}
\label{fig:dataset1}
\end{figure}

We present multiple types of pixel-wise image ground truths from different perspectives (Figure~\ref{fig:dataset1}). In particular, our dataset includes multiple geometry-related maps: a \underline{3D coordinate map} records the 3D coordinate information;
a \underline{normal map} records the normal vectors on the surface; a
\underline{depth map} records the distances from the object to the camera. 
Further, it also includes illumination-related masks: an \underline{albedo map} records the content of images without illumination information. Apart from these, a \underline{UV map}, which plays the role of indicators of textures and makes a bridge between geometry and texture space, and a \underline {background mask}, which indicates the location of film pixels as one, are also recorded. All diverse information helps us improve the effectiveness of the dewarping film task. Finally, a \underline{backward map}, which is used to restore the warped image in the last dewarping step, is calculated from the UV map.

Compared with scanning 3D shape of the real deformed objects, the physics computing engines simulate deformation in a completely automatic fashion. It is unnecessary to define any deformations applied to the real objects before capturing. By adjusting various parameters, the physics simulator can estimate the structured cortex set of true deformation situations without restriction. With diversity and randomness, the synthetic dataset instills the learned deep models with better generalization capability.

\subsection{Film Texture \& HDR Texture}
CQ500~\cite{chilamkurthy2018development} is a public, non-contrast head CT scan dataset containing 491 scans together with 193,317 slices, which is generously provided by the Centre for Advanced Research in Imaging, Neurosciences and Genomics (CARING), New Delhi, IN. We sample slices from a CT scan in CQ500 and typeset them as grids to generate film texture. Owing to the large variation in the number of slices in a scan, in most of cases, we select 16 and 24 slices and arrange 4*4 or 6*4 grids to represent different film texture shapes. We remove the bottom 10\% and top 10\% slices of a scan and sample randomly the desired number of slices with an equal interval for many times. Before typesetting, annotations including slice id, display field of view, scan name, slice thickness, and so on are marked on corners of slices by a software, sante dicom editor\footnote{Sante DICOM Editor: https://www.santesoft.com/}. We collect some indoor HDR textures originating from a hybrid of two sources: the Laval Indoor HDR dataset examples \cite{DBLP:journals/tog/GardnerSYSGGL17} and free pictures from HDR Haven\textsuperscript{\ref{hdrhaven}}. All these textures increase our dataset diversity and image variation.

\subsection{Deformation Simulating}
Regarding the hard accessibility of the medical films, we utilize the physics computing engines, Blender~\cite{blender}, to simulate the dynamic process of deformation. We apply soft-body physics characteristics with fixed volume to objects. The situations of the deformed film objects can be influenced by the soft-body attributes, the times of subdivision, the weights of vertices, and the force fields with different strengths and directions. In particular, the former two factors decide the hardness of the soft bodies. The weights of vertices define the force activation of different soft-body parts in force fields. The larger weights the vertices own, the more difficult the corresponding parts deform. Notably, we subdivide the film mesh into integer times of texture grids vertices and apply to subdivide modifier for rendering. The subdivide vertices contain the vertices of grids, which can be used to refine and output the exact coordinate. The dynamic deformation frames are captured in sequences and can be sampled randomly.

\subsection{Film Rendering}
\textbf{Configurations:} 
To increase the diversity of the dataset, we alter the configurations of camera, lighting, and texture in the rendering process and select a random frame from the entire dynamic deformation process. 
For each image, we apply the child-of constraints between the camera and captured objects. 
The relative position varies in a small range. We render 80\% of the images using lighting environments randomly sampled from the collection environment of HDR textures and the other 20\% with a simple backlight instead. We create a new material and adjust the transmission, IOR (Index of Refraction), roughness, specular, metallic, and so on to make film objects more realistic.

\textbf{Rich annotations:} 
For each image, we generate the 3D coordinate map, depth map, normal map, UV map, albedo map, and background mask. Considering the grid structure of the medical film, we also present the coordinates in world space or image space of the grid vertices. Besides, the camera lens and direction information also are included in this dataset.
    
\section{CT Film Recovery Network (FiReNet)}

\begin{figure*}[!t] 
\centerline{\includegraphics[width=0.85\linewidth]{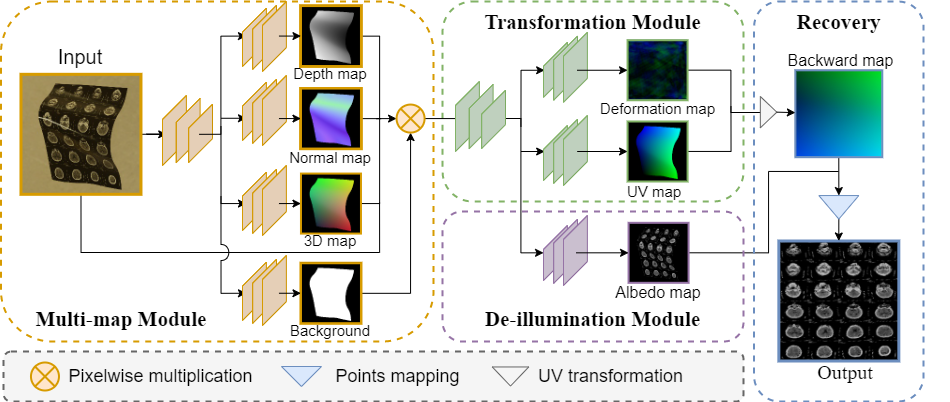}}
\caption{\textbf{Overview of the proposed network: FiReNet.}}
\label{fig:overview}
\end{figure*}

\subsection{Network Architecture}
As shown in Figure~\ref{fig:overview}, our network consists of three modules: \textit{multi-map module}, \textit{transformation module} and \textit{de-illumination module} and a post-processing operation: \textit{recovery}. In general, FireNet takes a warped CT film $\mathbf{I} \in \mathbb{R}^{h \times w \times 3}$ ($h$ and $w$ are image height and width, respectively) as input and predicts a UV map $\mathbf{M_{uv}} \in \mathbb{R}^{h \times w \times 2}$, where each point indicates its position on a texture map. To achieve the purpose of obtaining the texture map, we transform the UV map $\mathbf{M_{uv}}$, which maps textures to objects, to the backward map $\mathbf{B}  \in \mathbb{R}^{h \times w \times 2}$, which maps objects to textures by linear interpolation. 
In addition, we also generate the albedo map $\mathbf{M_{ab}} \in \mathbb{R}^{h \times w \times 1}$, where all illumination information is removed. Finally, we use bilinear sampling to sample the pixel values in $\mathbf{M_{ab}}$ to generate the final dewarped and de-illuminated CT film $\mathbf{D_{ab}} \in \mathbb{R}^{h \times w \times 1}$.

\textbf{Multi-map Module.}
We believe that the original image and its various annotations such as 3D coordinate map $\mathbf{M_{3D}} \in \mathbb{R}^{h \times w \times 3}$, normal map $\mathbf{M_{nor}}\in \mathbb{R}^{h \times w \times 3}$, and depth map $\mathbf{M_{dp}}\in \mathbb{R}^{h \times w \times 1}$ are closely related, hence unlike DewarpNet~\cite{das2019dewarpnet} using only 3D map, we use the information of three maps mentioned above by a shared encoder which extracts the features of the input image and 4 different decoders which generate $\mathbf{M_{3D}}$, $\mathbf{M_{nor}}$, $\mathbf{M_{dp}}$ and background map $\mathbf{M_{bg}} \in \mathbb{R}^{h \times w \times 1}$ respectively, with U-Net~\cite{ronneberger2015u} as the backbone network. When these maps are generated, they are combined and filtered by $\mathbf{M_{bg}}$ and sent to the subsequent modules for more processing.


\textbf{Transformation Module.}
Compared with backward map $\mathbf{B}$ generated directly from a model~\cite{das2019dewarpnet}, we design the transformation module to generate the UV map $\mathbf{M_{uv}}$ which is pixel-wise aligned with the original image and is stable and robust for restoration (refer to Section~\ref{section:uvbw_study}), but some regions especially around the edges are often missed because the number of valid points in $\mathbf{M_{uv}}$ is less than that in the backward map $\mathbf{B}$ and most of the valid points in $\mathbf{M_{uv}}$ are concentrated in the middle part. Therefore, in the transformation module for geometry refinement, we also generate a \underline{deformation map} $\mathbf{M_{df}}  \in \mathbb{R}^{h \times w \times 2}$, which is another form of the backward map and represents the relative movement position of each point from the input image to the unwarped image, and we utilize $\mathbf{M_{df}}$ to enhance $\mathbf{M_{uv}}$. Finally, we will use interpolation to make the final image. We also invoke the U-Net as the basic framework and take the integrated features from the previous module as inputs, and then use two different decoders to generate the corresponding UV map $\mathbf{M_{uv}}$ and deformation map $\mathbf{M_{df}}$, both of which have the same size as the original image.
Afterwards, we combine the UV map $\mathbf{M_{uv}}$ and deformation map $\mathbf{M_{df}}$ by converting the deformation map to UV map $\mathbf{M^{'}_{uv}}$ where points are used to fill blanks in $\mathbf{M_{uv}}$, which is finally converted to backward map $\mathbf{B}$ for CT film recovery.

\textbf{De-illumination Module.}
To remove the disturbing from lights, we design the de-illumination module where an additional decoder is used to generate the albedo map $\mathbf{M_{ab}}$, which represents the image with the lighting information removed. Finally, in the recovery stage, we use $\mathbf{B}$ to restore the dewarped albedo map $\mathbf{D_{ab}}$, which becomes the final output. 

\textbf{Recovery Details.} 
There are two steps for film recovery: 

(1) UV transformation (from the UV map and deformation map to backward map): For the UV map $\mathbf{M_{uv}} \in \mathbb{R}^{h \times w \times 2}$, we have its coordinate matrix $\mathbf{C} \in \mathbb{R}^{h \times w \times 2}$, where each element refers to the coordinate of its corresponding point in $\mathbf{M_{uv}}$ ($\forall c_{ij} \in \mathbf{C}, i\in [0,1,\cdots,h-1], j\in[0,1,\cdots,w-1]$, where $w$ and $h$ are width and height of $\mathbf{M_{uv}}$, respectively),
and value matrix $\mathbf{V}$, where each element stores the values of the corresponding point in $\mathbf{M_{uv}}$ ($\mathbf{V} = \mathbf{M_{uv}}$). After filtering out points with zero values on $\mathbf{V}$, we consider 
$\mathbf{V}$ as the coordinate matrix $\mathbf{C'}$ of the backward map $\mathbf{B}$ and $\mathbf{C}$ as the value matrix $\mathbf{V'}$ of backward map $\mathbf{B}$. 
For the deformation map $\mathbf{M_{df}}$, we again have its coordinate matrix $\mathbf{C_{df}}$ and value matrix $\mathbf{V_{df}}$, and select the 4 corners and 3 points for each edge and mix them into $\mathbf{C'}$ and $\mathbf{V'}$.
Finally, we interpolate $\mathbf{V'}$ by $\mathbf{C'}$ to get backward map $\mathbf{B}$. 

(2) Points mapping (from the backward map to dewarped film): For the backward map, we have its value matrix where the value of each point indicates the coordinate of the original warped film. Therefore, we can use bilinear sampling to sample the pixel value in the input image to generate unwarped film or sample pixel value in $\mathbf{M_{ab}}$ to generate dewarped and de-illuminated film $\mathbf{D_{ab}}$.

\subsection{Training Loss Function}
In our training, we perform the end-to-end training on our FiReNet with multiple objectives. In the multi-map module, we train the four maps with \textit{$L_1$ loss}. Taking the 3D coordinate map as an example:
\begin{equation} \label{eq:1}
    L_{3D} = || \mathbf{\hat{M}_{3D}} - \mathbf{M_{3D}} ||_{1},
\end{equation}
where $\mathbf{\hat{M}_{3D}}$ represents the predicted 3D coordinate map, $\mathbf{M_{3D}}$ is the ground truth, and $L_{3D}$ represents the loss value on $\mathbf{M_{3D}}$. Similarly, We define other losses of $L_{bg}$, $L_{nor}$, and $L_{dp}$. Therefore, the total loss for shape module, $L_{shape}$, is formulated as follows:
\begin{equation}
L_{shape} = L_{3D} + L_{nor} + L_{dp} + L_{bg}
\end{equation}

In transformation module, we first perform the same operations as before on the maps of $\mathbf{M_{ab}}$, $\mathbf{M_{uv}}$, and $\mathbf{M_{df}}$ to get the corresponding losses, which are denoted as $L_{ab}$, $L_{uv}$, and $L_{df}$, respectively. In addition, we also add a \textit{shifting loss} $L_{shift}$ to reduce shifting, a \textit{disturbance loss} $L_{disturb}$ to reduce total variance, and a \textit{de-shifted difference loss} $L_{diff}$ to reduce variance of each point of $\mathbf{M_{df}}$. Denoting the predicted deformation map by $\mathbf{\hat{M}_{df}}$ and the ground truth by $\mathbf{M_{df}}$, we represent the entire process as follows:
\begin{equation}
    \begin{split}
&L_{shift}  = ||\mathbf{\mu}||_{1}, ~L_{disturb}  = ||\mathbf{\sigma}||_{1}, \\
&L^{'}_{diff}  = min(||\mathbf{\Delta M_{df}}||_{1} , ||\mathbf{\Delta M_{df}} - \mathbf{\mu}||_{1})), \\
&L_{diff} = \begin{cases}
L^{'}_{diff}&\mathbf{\Delta M_{df}} \odot (\mathbf{\Delta M_{df}} - \mu) > 0, \\
0 &\mathbf{\Delta M_{df}} \odot (\mathbf{\Delta M_{df}} - \mu) <= 0. \\
          \end{cases}
    \end{split}
    \label{eq:transLoss}
\end{equation}
where $\mathbf{\Delta M_{df}}  = \mathbf{\hat{M}_{df}} - \mathbf{M_{df}}$; $\mathbf{\mu} = \mu(\mathbf{\Delta M_{df}})$; $\mathbf{\sigma} = \sigma(\mathbf{\Delta M_{df}})$.

For deformation map, we hope our model to focus on relative deformation, hence in (\ref{eq:transLoss}) we first calculate the difference $\mathbf{\Delta M_{df}}$ between the predicted deformation map $\mathbf{\hat{M}_{df}}$ and ground truth $\mathbf{M_{df}}$, and then calculate the mean $\mu$ and the variance $\sigma$ of $\mathbf{\Delta M_{df}}$. To focus on the relative image warping, we separate the shift loss $L_{shift}$ from $\mathbf{\Delta M_{df}}$, and add standard variation as disturbing loss $L_{disturb}$. In particular, we hope the prediction $\mathbf{\hat{M}_{df}}$ to approximate the relative location if far away from both relative and the absolute location, but we should not affect points already close to the absolute location, hence we calculate $L_{diff}$ as mentioned above.
Therefore, the total loss of the deformation map $L_{trans}$ is:
\begin{equation} \label{eq:loss_trans}
    L_{df} = L_{shift} + \alpha * L_{disturb} + \beta * L_{diff}, 
\end{equation}
where $\alpha$ and $\beta$ are weights for $L_{disturb}$ and $L_{diff}$.
Similarly, we can get $L_{uv}$ for UV map according to $L_{df}$, and $L_{ab}$ according to $L_{3D}$. The loss for transformation module can be represented as follows: 
\begin{equation}
    L_{trans} = L_{df} + L_{uv} + L_{ab}.
\end{equation}

Finally, the total loss for training the complete network is given as : $L = L_{shape} + L_{trans}$.

\section{Experiments}
We evaluate our approach with several experiments on the test dataset of approximately 2000 images.
Because the material and properties of CT Flim are different from the document and there are few folds and the deformation of CT film is smoother, all experiments are based on CT film.
We train DewarpNet \cite{das2019dewarpnet} on our dataset for comparison, and we also analyze some problems of the previous method directly generating backward maps, and finally, we provide an ablation study to show and explain the performance of our modules, including multi-map, deformation modules. Besides, we provide a visual evaluation of hard examples and real photos shown in Figure~\ref{fig:hard}.



\subsection{Setup}
\textbf{Evaluation Metrics.} 
To evaluate the similarity between the original and the dewarped CT film image, we choose PSNR, SSIM~\cite{wang2004image}, and MS-SSIM~\cite{wang2003multiscale} as our evaluation metrics. Peak signal-to-noise ratio (PSNR), is the ratio between the maximum possible power of a signal and the power of corrupting noise that affects the fidelity of its representation and is often used in image quality evaluation. SSIM computes the similarity of the mean pixel value and variance within each image patch and averages over all patches in an image. MS-SSIM applies SSIM at multiple scales using a Gaussian pyramid, better suited for evaluation of global similarity between results and ground truth. In these experiments, to avoid the effects of image shifting, we estimate the de-shifted results on these metrics. 

\textbf{Training Details.}
For training, firstly, input images are resized to $256\times256$ and we set Adam~\cite{DBLP:journals/corr/KingmaB14} as the optimizer, learning rate 0.001, batch size 16 and set $\alpha$ to 2 and $\beta$ to 2 in (\ref{eq:loss_trans}). All maps and images are linearly scaled to the range $[-1,1]$.

\subsection{Comparative Experiment} \label{section:comparison}
\begin{figure}[t] 
    \begin{center}
      \includegraphics[width=\linewidth]{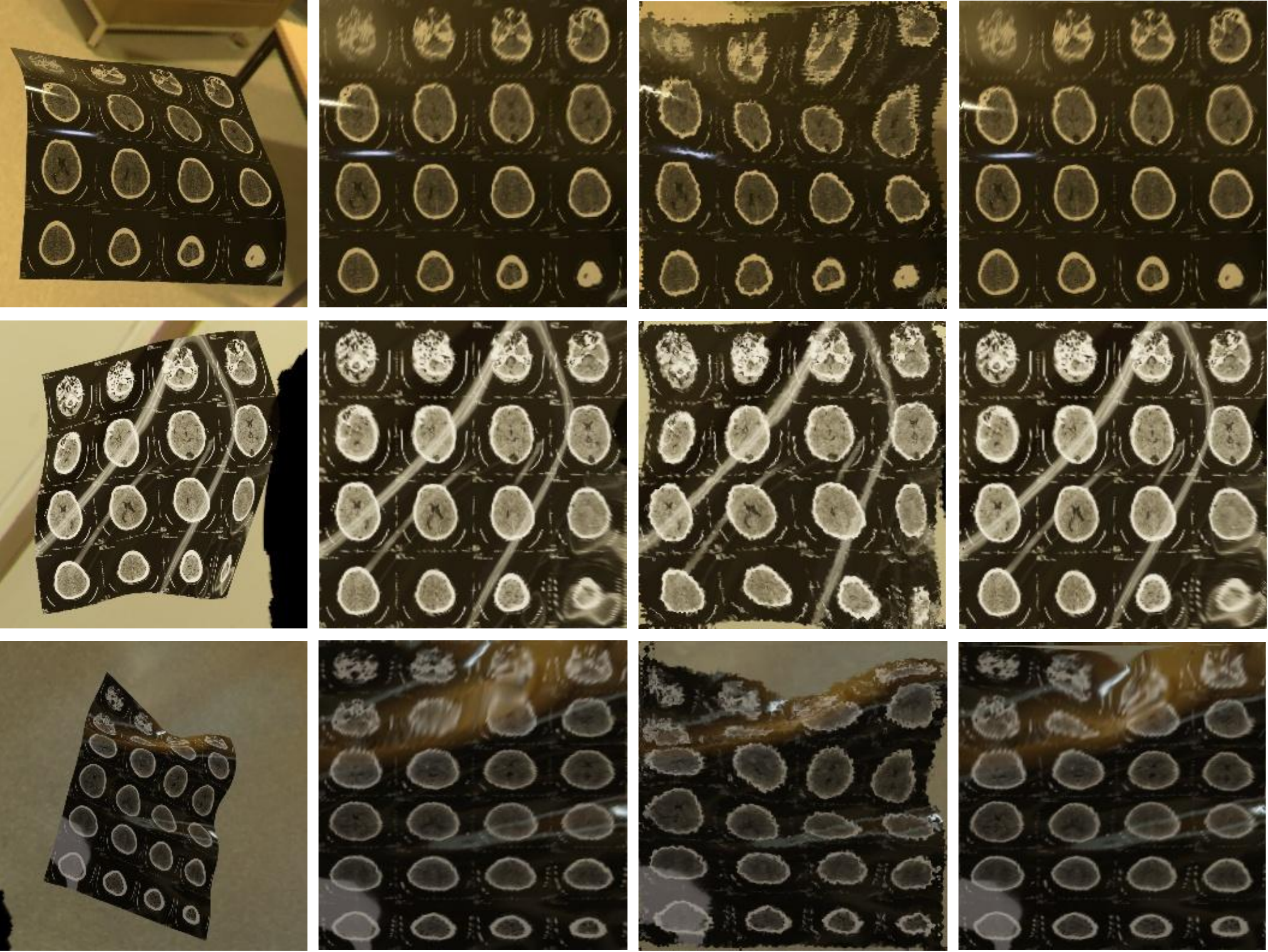}
    \end{center}
       \caption{\textbf{Qualitative comparison between our approach and others.} The 1st column: warped  CT films. The 2nd column: ground truth. The 3rd column: DewarpNet~\cite{das2019dewarpnet}. The 4th column: ours. Our results look better.}
       \label{fig:visual_comparison}
    \end{figure}

We test DewarpNet \cite{das2019dewarpnet} and our method on our CTFilm20k dataset.
DewarpNet directly generates a backward map $\mathbf{B}$, while we generate a UV map, which is more robust for film restoration.
Results in Table~\ref{table:compare} and Figure~\ref{fig:visual_comparison} show our advantage over other methods. The improvement on the performance of evaluation with de-shifted backward map $\mathbf{B}$ in Table \ref{table:compare} confirms the benefit of using deformation module and de-shifted difference loss, and the overall improvement is mainly due to the use of the UV map and network architecture.

\begin{table} 
    \begin{center}
    \begin{tabular}{lccc}
    \multicolumn{4}{c}{Evaluation with $\mathbf{B}$} \\
    \hline
    Methods & PSNR $\uparrow$  & SSIM $\uparrow$ & MS-SSIM $\uparrow$  \\
    \hline
    DewarpNet~\cite{das2019dewarpnet} &16.98  &0.4501 &0.6879\\
    Ours & 25.30 & 0.8621 & 0.9523\\
    \hline
    \multicolumn{4}{c}{Evaluation with de-shifted $\mathbf{B}$} \\
    \hline
    Methods & PSNR $\uparrow$ & SSIM $\uparrow$ & MS-SSIM $\uparrow$  \\
    \hline
    DewarpNet~\cite{das2019dewarpnet} &17.18 &0.4611 &0.6932\\
    Ours & 25.60 & 0.8715 & 0.9564\\
     \hline
    \end{tabular}
    \end{center}
    \caption{\textbf{Quantitative comparison between our method and  DewarpNet~\cite{das2019dewarpnet}.} Methods are evaluated by PSNR, SSIM and MS-SSIM. For all of these metrics, higher value means better. The upper table is the evaluation on dewarped films with normal backward map $\mathbf{B}$, and the lower with de-shifted $\mathbf{B}$ where "de-shifted" means subtracting the average of difference between the prediction and ground truth. Our method performs better than DewarpNet~\cite{das2019dewarpnet}.}
\label{table:compare}
\end{table}

\subsection{Study of UV and Backward Maps} \label{section:uvbw_study}
During our research, it is found that, compared with directly generated backward mapping, the UV map has the following advantages: the UV map is pixel-wise aligned with the original image, while the backward map is not, which leads to more complexity for network training and less reliability. To prove our assumption, we design a comparison experiment of the performance of the backward map and the UV map to explore the visual problems of the dewarping network. We use simple U-Net without tricks to construct two identical transformation networks: one is from transform 3D maps to backward maps, which we denote as $Net_{bw}$ (this is what has been done in DewarpNet), and the other is 3D maps to UV maps which we denoted as $Net_{uv}$. All experiment parameters are exactly the same and both networks are trained for 50 epochs to reach the final model and evaluated on the same test dataset as previous experiments. 
In comparison, we compare the backward maps (which the UV map will be converted into a backward map for comparison) and the dewarped image results. Results are shown on Table \ref{uvbw},

\begin{table} 
\begin{center}
\begin{tabular}{lccc}
Map & PSNR $\uparrow$  & SSIM $\uparrow$ & MS-SSIM $\uparrow$\\ 
\hline
$\mathbf{B_{bw}}$     & 44.79 & 0.9994 & 0.9985\\ 
$\mathbf{B_{uv}}$ & 10.77 & 0.8038 & 0.4894 \\
$\mathbf{D_{bw}}$ & 18.46 & 0.6126 & 0.8044 \\
$\mathbf{D_{uv}}$ & 15.13 & 0.2642 & 0.5028 \\
$\mathbf{\Tilde{D}_{bw}}$ & 19.53 & 0.6172 & 0.8088  \\
$\mathbf{\Tilde{D}_{uv}}$ & 22.83 & 0.7791 & 0.9198 \\
\hline
\end{tabular}
\end{center}
\caption{\textbf{Quantitative comparison of UV and backward map.} $\mathbf{B_{bw}}$ refers to the backward map predicted by $Net_{bw}$; $\mathbf{B_{uv}}$ refers to the backward map generated by UV map from $Net_{uv}$; $\mathbf{D_{bw}}$ refers to the dewarped image generated by $\mathbf{B_{bw}}$; $\mathbf{D_{uv}}$ refers to the dewarped image generated by $\mathbf{B_{uv}}$; $\mathbf{\Tilde{D}_{bw}}$ refers to the dewarped image generated by de-shifted $\mathbf{B}$; $\mathbf{\Tilde{D}_{uv}}$ refers to the dewarped image generated by de-shifted $\mathbf{B_{uv}}$.}
\label{uvbw}
\end{table}

\begin{figure}[t] 
\begin{center}
   \includegraphics[width=\linewidth]{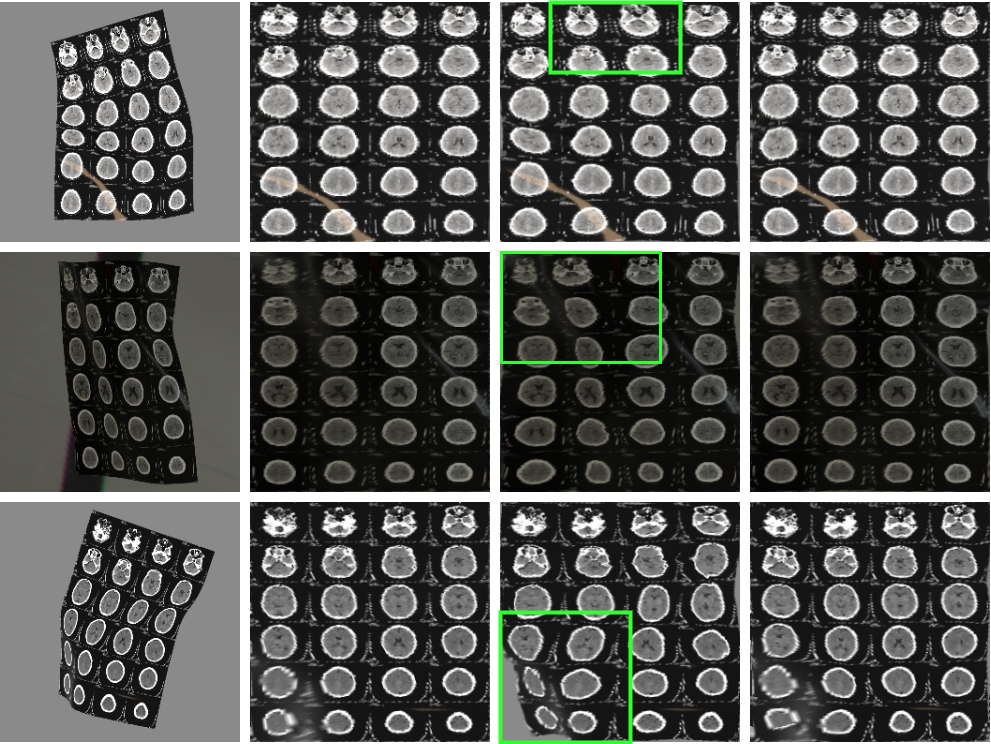}
\end{center}
   \caption{\textbf{Visual comparison for UV and backward maps.} The 1st column: warped CT films. The 2nd column: ground truth. The 3rd column: dewarped image using backward map. The 4th column: dewarped image using UV map. Green boxes: regions with noticeable deformations.}
   \label{fig:uvbw}
\end{figure}

Per Table~\ref{uvbw}, we can clearly find that the PSNR and SSIM scores by $\mathbf{B_{bw}}$ (or $\mathbf{D_{bw}}$) is numerically better than those by $\mathbf{B_{uv}}$ (or $\mathbf{B_{uv}}$) , but visually, the images generated by using the UV map are much closer to our expectations (Figure~\ref{fig:uvbw}). Therefore, 
here these numerical indicators do not indicate the visual restoration performance. In our experiments, we find the reason for this phenomenon: In Figure~\ref{fig:uvbw}, which shows good visualization of the results generated by the UV map, there is a shift in the predicted UV map for some degree with respect to the original image, which causes the pixel-wise numerical indicators to deteriorate. 
Therefore, the pixel-wise indicators may be somewhere unsuitable and we proposed to augment the original losses with two extra loss functions: \textit{shifting loss} and \textit{de-shifted difference loss}. With the help of the extra losses, we obtain $\mathbf{\Tilde{D}_{bw}}$ and $\mathbf{\Tilde{D}_{uv}}$: now the quantitative order matches with the order of visual quality.

\subsection{Ablation Study}
To further analyze our method, we conduct an ablation study to characterize the roles of each module. Results are shown in Table \ref{table:ablation}.

\begin{table}
\begin{center}
\begin{tabular}{ccccc}
\multicolumn{5}{c}{Evaluation with de-shifted $\mathbf{B}$} \\
\hline
UV & MM  & Deform & PSNR $\uparrow$ & MS-SSIM $\uparrow$\\
\hline
\checkmark & & &  25.16& 0.9515\\
\checkmark & \checkmark & & 25.51 & 0.9553 \\
\checkmark & \checkmark & \checkmark & 25.60 & 0.9564 \\
\hline
baseline &&& 23.76 & 0.9446\\
\end{tabular}
\end{center}
\caption{\textbf{Ablation study.} UV: using UV maps to dewarp films. MM: generating multiple maps. Deform: adding deformation module.}
\label{table:ablation}
\end{table}

\textbf{UV map}: UV map is the key part of our method to restore the image. We compare our model which generates UV maps for restoration, with the baseline model which uses backward maps for restoration. According to Table~\ref{table:ablation}, we can find that UV map outperforms the backward map (25.16dB vs 23.76dB in PSNR).

\textbf{Multi-map}: 
We study the comparison between using only a 3D coordinate map and using multiple maps that include 3D coordinate, normal, and depth maps. 
From Table \ref{table:ablation}, we can find an improvement on these metrics (from 25.16dB to 25.51dB in PSNR).

\textbf{Deformation module}: We compare the performance of networks with and without the deformation module. The performance is improved from 25.51dB to 25.60dB in PSNR.
                                                              
\begin{figure}[t] 
    \begin{center}
       \includegraphics[width=\linewidth]{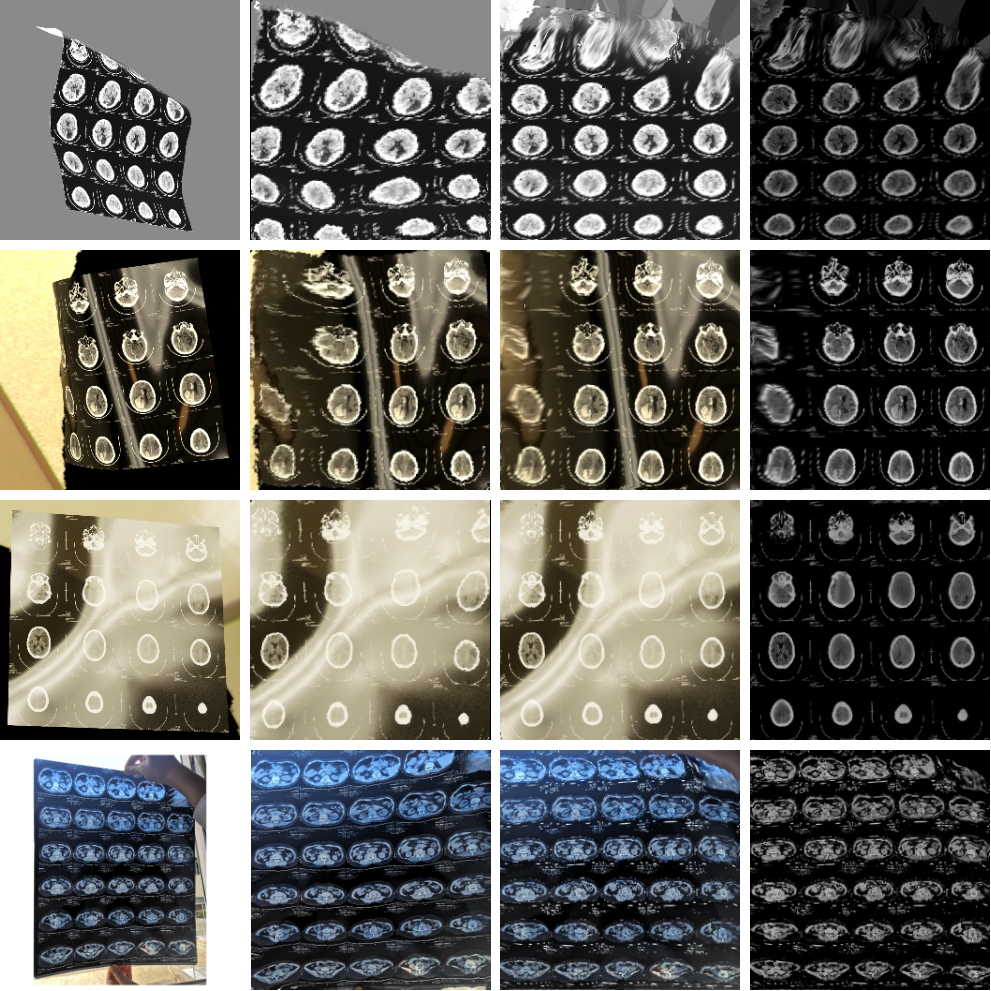}
    \end{center}
       \caption{\textbf{Hard examples and a real film.} The 1st, 2nd and 3rd rows: hard examples of warped CT films. The 4th row: real warped films. The 1st column: warped films. The 2nd column: results from DewarpNet. The 3rd and 4th columns: results from our method. Our method can recover the existing information as much as possible.}
       \label{fig:hard}
    \end{figure}
    
\subsection{Robust Evaluation}
For robust evaluation, we show our results on hard examples in Figure~\ref{fig:hard}, where the first and the second examples are over-warped but rectified in a certain degree by our approach, and the third one filled with strong lights are rectified as much as possible. In contrast, the results from DewarpNet exhibit more local deformations. The 4th image is captured from a real mobile camera. Furthermore, the 4th image contains a chest and does not belong to the head and neck region, hence deviating largely from our simulated training images. Also, this image features an occlusion by the hand in the top right corner of the film. Yet our recovery is rather reasonable and the grids are rectified.

According to these results, we can observe that our approach performs robustly against strong lights or variations and has some generalization capability from simulated to real scenarios. 



\section{Conclusion and Future Work}
In this work, we proposed our FiReNet, a novel dewarping network for CT film recovery. Our approach is robust to film content, reflection, and background and is insensitive to slight image shifting. Through extensive experiments, we validated the advantages of our approach, which outperforms the previous approaches. Moreover, we contribute the CTFilm20K dataset, which is the first large dataset for CT film dewarping, and full of various types of maps and annotations. However, there still exist some limitations in our work. For our dataset, the types of CT film is limited, most of them are images of brain and bones. Also, the images are simulated and none of them is captured directly from real-world scenarios. We will collect more data and validate our method to real scenarios.





{\small
\bibliographystyle{ieee_fullname}
\bibliography{egbib}

\begin{thebibliography}{10}\itemsep=-1pt

\bibitem{brown2001document}
Michael~S Brown and W~Brent Seales.
\newblock Document restoration using 3d shape: a general deskewing algorithm
  for arbitrarily warped documents.
\newblock In {\em ICCV}. IEEE, 2001.

\bibitem{cao2003cylindrical}
Huaigu Cao, Xiaoqing Ding, and Changsong Liu.
\newblock A cylindrical surface model to rectify the bound document image.
\newblock In {\em ICCV}. IEEE, 2003.

\bibitem{chilamkurthy2018development}
Sasank Chilamkurthy, Rohit Ghosh, Swetha Tanamala, Mustafa Biviji, Norbert~G
  Campeau, Vasantha~Kumar Venugopal, Vidur Mahajan, Pooja Rao, and Prashant
  Warier.
\newblock Development and validation of deep learning algorithms for detection
  of critical findings in head ct scans.
\newblock {\em arXiv preprint arXiv:1803.05854}, 2018.

\bibitem{blender}
Blender~Online Community.
\newblock {\em Blender - a 3D modelling and rendering package}.
\newblock Blender Foundation, Stichting Blender Foundation, Amsterdam, 2018.

\bibitem{courteille2007shape}
Fr{\'e}d{\'e}ric Courteille, Alain Crouzil, Jean-Denis Durou, and Pierre
  Gurdjos.
\newblock Shape from shading for the digitization of curved documents.
\newblock {\em Machine Vision and Applications}, 18(5):301--316, 2007.

\bibitem{das2019dewarpnet}
Sagnik Das, Ke Ma, Zhixin Shu, Dimitris Samaras, and Roy Shilkrot.
\newblock Dewarpnet: Single-image document unwarping with stacked 3d and 2d
  regression networks.
\newblock In {\em CVPR}. IEEE, 2019.

\bibitem{das2017common}
Sagnik Das, Gaurav Mishra, Akshay Sudharshana, and Roy Shilkrot.
\newblock The common fold: utilizing the four-fold to dewarp printed documents
  from a single image.
\newblock In {\em Proceedings of the 2017 ACM Symposium on Document
  Engineering}, pages 125--128, 2017.

\bibitem{ezaki2005dewarping}
Hironori Ezaki, Seiichi Uchida, Akira Asano, and Hiroaki Sakoe.
\newblock Dewarping of document image by global optimization.
\newblock In {\em Eighth International Conference on Document Analysis and
  Recognition (ICDAR'05)}, pages 302--306. IEEE, 2005.

\bibitem{forsyth2001shape}
David~A Forsyth.
\newblock Shape from texture and integrability.
\newblock In {\em ICCV}. IEEE, 2001.

\bibitem{DBLP:journals/tog/GardnerSYSGGL17}
Marc{-}Andr{\'{e}} Gardner, Kalyan Sunkavalli, Ersin Yumer, Xiaohui Shen,
  Emiliano Gambaretto, Christian Gagn{\'{e}}, and Jean{-}Fran{\c{c}}ois
  Lalonde.
\newblock Learning to predict indoor illumination from a single image.
\newblock {\em {ACM} Trans. Graph.}, 36(6):176:1--176:14, 2017.

\bibitem{kim2015document}
Beom~Su Kim, Hyung~Il Koo, and Nam~Ik Cho.
\newblock Document dewarping via text-line based optimization.
\newblock {\em Pattern Recognition}, 48(11):3600--3614, 2015.

\bibitem{DBLP:journals/corr/KingmaB14}
Diederik~P. Kingma and Jimmy Ba.
\newblock Adam: {A} method for stochastic optimization.
\newblock In {\em ICLR}, 2015.

\bibitem{koo2009composition}
Hyung~Il Koo, Jinho Kim, and Nam~Ik Cho.
\newblock Composition of a dewarped and enhanced document image from two view
  images.
\newblock {\em IEEE Transactions on Image Processing}, 18(7):1551--1562, 2009.

\bibitem{liang2008geometric}
Jian Liang, Daniel DeMenthon, and David Doermann.
\newblock Geometric rectification of camera-captured document images.
\newblock {\em IEEE Transactions on Pattern Analysis and Machine Intelligence},
  30(4):591--605, 2008.

\bibitem{liu2015restoring}
Changsong Liu, Yu Zhang, Baokang Wang, and Xiaoqing Ding.
\newblock Restoring camera-captured distorted document images.
\newblock {\em International Journal on Document Analysis and Recognition
  (IJDAR)}, 18(2):111--124, 2015.

\bibitem{lu2006document}
Shijian Lu and Chew~Lim Tan.
\newblock Document flattening through grid modeling and regularization.
\newblock In {\em 18th International Conference on Pattern Recognition
  (ICPR'06)}. IEEE, 2006.

\bibitem{ma2018docunet}
Ke Ma, Zhixin Shu, Xue Bai, Jue Wang, and Dimitris Samaras.
\newblock Docunet: document image unwarping via a stacked u-net.
\newblock In {\em CVPR}. IEEE, 2018.

\bibitem{malik1997computing}
Jitendra Malik and Ruth Rosenholtz.
\newblock Computing local surface orientation and shape from texture for curved
  surfaces.
\newblock {\em International journal of computer vision}, 23(2):149--168, 1997.

\bibitem{meng2018exploiting}
Gaofeng Meng, Yuanqi Su, Ying Wu, Shiming Xiang, and Chunhong Pan.
\newblock Exploiting vector fields for geometric rectification of distorted
  document images.
\newblock In {\em ECCV}, 2018.

\bibitem{meng2014active}
Gaofeng Meng, Ying Wang, Shenquan Qu, Shiming Xiang, and Chunhong Pan.
\newblock Active flattening of curved document images via two structured beams.
\newblock In {\em CVPR}, 2014.

\bibitem{ostlund2012laplacian}
Jonas {\"O}stlund, Aydin Varol, Dat~Tien Ngo, and Pascal Fua.
\newblock Laplacian meshes for monocular 3d shape recovery.
\newblock In {\em ECCV}. Springer, 2012.

\bibitem{ronneberger2015u}
Olaf Ronneberger, Philipp Fischer, and Thomas Brox.
\newblock U-net: Convolutional networks for biomedical image segmentation.
\newblock In {\em MICCAI}. Springer, 2015.

\bibitem{tian2011rectification}
Yuandong Tian and Srinivasa~G Narasimhan.
\newblock Rectification and 3d reconstruction of curved document images.
\newblock In {\em CVPR}. IEEE, 2011.

\bibitem{tsoi2007multi}
Yau-Chat Tsoi and Michael~S Brown.
\newblock Multi-view document rectification using boundary.
\newblock In {\em CVPR}. IEEE, 2007.

\bibitem{ulges2004document}
Adrian Ulges, Christoph~H Lampert, and Thomas Breuel.
\newblock Document capture using stereo vision.
\newblock In {\em Proceedings of the 2004 ACM symposium on Document
  engineering}, pages 198--200, 2004.

\bibitem{ulges2005document}
Adrian Ulges, Christoph~H Lampert, and Thomas~M Breuel.
\newblock Document image dewarping using robust estimation of curled text
  lines.
\newblock In {\em Eighth International Conference on Document Analysis and
  Recognition (ICDAR'05)}, pages 1001--1005. IEEE, 2005.

\bibitem{wada1997shape}
Toshikazu Wada, Hiroyuki Ukida, and Takashi Matsuyama.
\newblock Shape from shading with interreflections under a proximal light
  source: Distortion-free copying of an unfolded book.
\newblock {\em International Journal of Computer Vision}, 24(2):125--135, 1997.

\bibitem{wang2004image}
Zhou Wang, Alan~C Bovik, Hamid~R Sheikh, and Eero~P Simoncelli.
\newblock Image quality assessment: from error visibility to structural
  similarity.
\newblock {\em IEEE transactions on image processing}, 13(4):600--612, 2004.

\bibitem{wang2003multiscale}
Zhou Wang, Eero~P Simoncelli, and Alan~C Bovik.
\newblock Multiscale structural similarity for image quality assessment.
\newblock In {\em The Thrity-Seventh Asilomar Conference on Signals, Systems \&
  Computers, 2003}, volume~2, pages 1398--1402. Ieee, 2003.

\bibitem{witkin1981recovering}
Andrew~P Witkin.
\newblock Recovering surface shape and orientation from texture.
\newblock {\em Artificial intelligence}, 17(1-3):17--45, 1981.

\bibitem{yamashita2004shape}
Atsushi Yamashita, Atsushi Kawarago, Toru Kaneko, and Kenjiro~T Miura.
\newblock Shape reconstruction and image restoration for non-flat surfaces of
  documents with a stereo vision system.
\newblock In {\em Proceedings of the 17th International Conference on Pattern
  Recognition, 2004. ICPR 2004.}, volume~1, pages 482--485. IEEE, 2004.

\bibitem{you2017multiview}
Shaodi You, Yasuyuki Matsushita, Sudipta Sinha, Yusuke Bou, and Katsushi
  Ikeuchi.
\newblock Multiview rectification of folded documents.
\newblock {\em IEEE transactions on pattern analysis and machine intelligence},
  40(2):505--511, 2017.

\bibitem{zhang2009unified}
Li Zhang, Andy~M Yip, Michael~S Brown, and Chew~Lim Tan.
\newblock A unified framework for document restoration using inpainting and
  shape-from-shading.
\newblock {\em Pattern Recognition}, 42(11):2961--2978, 2009.

\bibitem{zhang2008improved}
Li Zhang, Yu Zhang, and Chew Tan.
\newblock An improved physically-based method for geometric restoration of
  distorted document images.
\newblock {\em IEEE Transactions on Pattern Analysis and Machine Intelligence},
  30(4):728--734, 2008.

\bibitem{zhou2019handbook}
S~Kevin Zhou, Daniel Rueckert, and Gabor Fichtinger.
\newblock {\em Handbook of medical image computing and computer assisted
  intervention}.
\newblock Academic Press, 2019.

\end{thebibliography}
}

\end{document}